\documentclass{article}

\usepackage{xcolor}

\PassOptionsToPackage{numbers,sort&compress}{natbib}
\usepackage{PRIMEarxiv}
\usepackage[utf8]{inputenc}
\usepackage[T1]{fontenc}
\usepackage{natbib}
\usepackage{hyperref}
\usepackage{url}
\usepackage{booktabs}
\usepackage{nicefrac}
\usepackage{microtype}
\usepackage{fancyhdr}
\usepackage{float}
\usepackage{listings}
\usepackage[font=small,labelfont=bf]{caption}
\usepackage{enumitem}
\usepackage{pdfpages}
\usepackage{amsmath,amsfonts,amssymb,amsthm,mathtools}
\usepackage{dsfont}
\usepackage{bm}

\usepackage{amsfonts}       % blackboard math symbols

\usepackage{graphicx}
\graphicspath{{./}{../}}
\usepackage{tikz}
\usepackage{tikz-cd}
\usepackage{pgfplots}
\pgfplotsset{compat=1.18}

\usepackage[utf8]{inputenc} % allow utf-8 input
\usepackage[T1]{fontenc}    % use 8-bit T1 fonts

\usepackage[nameinlink,noabbrev]{cleveref}

\theoremstyle{definition}

\theoremstyle{plain}

\theoremstyle{remark}

\title{
FormalGen: Agentic formalization and verification of physical mechanisms for compositional  design

}

\author{
\textbf{Lee Marom}$^{1}$ \quad
\textbf{Markus J. Buehler}$^{1,2,3,*}$ \\[0.6em]
$^{1}$Department of Civil and Environmental Engineering,\\
$^{2}$Department of Mechanical Engineering,\\
$^{3}$Center for Computational Science and Engineering,\\
Schwarzman College of Computing,\\
Massachusetts Institute of Technology, Cambridge, MA 02139, USA
\\[0.4em]
$^{*}$Corresponding author:
\texttt{mbuehler@mit.edu}
}

\date{}

\begin{document}

\maketitle
\thispagestyle{empty}

\begin{abstract}
Artificial intelligence (AI) is undergoing a transition from open-ended to formal mathematical reasoning. Can a similar transition be achieved in the physical sciences, where discovery also requires constructing, composing and testing representations of physical reality? Here we introduce FormalGen, a multi-agent framework that brings this paradigm to physics-driven design, applied to multiscale materials. Agents formalize published mechanisms as category-theoretic models, propose mutations with genetic operators, and use Lean to check behavioral claims before finite-element testing. In one mutation, agents couple sunflower phototropism to bamboo graded-wall mechanics through bending-induced fiber reorientation. The model predicts a 2.46-fold increase in flexural rigidity, and simulations yield 2.45-fold between prescribed orientations. Ranking by interface and domain compatibility finds 100 verified connections with 97\% fewer candidates than connectivity-based ranking. From 8 to 32 mechanisms, proofs of individual laws increase fourfold while two-mechanism chains certified by proof reuse increase 27-fold, a basis for cumulative discovery.
\end{abstract}

\section*{Introduction}
\label{sec:introduction}

Artificial intelligence (AI) is undergoing a fundamental transition from open-ended reasoning to formal mathematical reasoning, while developing emerging capabilities in solving complex scientific problems~\cite{wang2023science,buehler2024,ghafarollahi2025,boiko2023,gottweis2026,miao2026,mathlib2020,loo_agentic_2026,huang_cascade_2025,song_llm_2025}. The recent AI-assisted proof of the Riemann hypothesis~\cite{alpoge2026remark,lean2026riemann}, independently verified in Lean~\cite{demoura2021}, demonstrates the extraordinary potential of combining generative exploration with formal mathematical systems. Can a similar transition be achieved in the physical sciences, where discovery requires not only establishing mathematical correctness but also constructing, composing and testing representations of physical reality?

To illustrate the complexity of scientific discovery, consider this example. In a cloud, water molecules assemble into a branching snow crystal~\cite{libbrecht2017}. Below, meltwater seeps through soil pores and rises through trees before returning to the air~\cite{savage2010,stroock2014}. Rooted in that soil, a young plant bends toward light as cells on its shaded side elongate faster~\cite{moulton2020}. Across these systems, the response of the whole depends on how local processes interact. A multiscale model seeks to explain how those interactions give rise to an observed response~\cite{wegst2015,marom2025,marom2026}. Design poses the corresponding question in reverse. Given a desired response, which processes should be connected, and how should they interact across scales?

AI agents can extract relations from scientific papers, propose hypotheses, and coordinate computational tools~\cite{wang2023science,buehler2024,ghafarollahi2025,boiko2023,gottweis2026,miao2026}. In mechanics, agents have coupled this reasoning to code generation and numerical solution~\cite{ni2024}. Knowledge graphs can identify relationships among mechanisms described in different sources~\cite{hogan2021,buehler2024,ghafarollahi2025,yang2026prospector}, but this reasoning is open-ended. It ranges across disciplines and proposes connections by analogy, and it leaves open whether a proposed connection holds. To use a proposed connection in a calculation, we need to know which physical quantity passes between the laws and under what conditions the connection is valid.

Formal libraries make machine-checked definitions and results available for reuse~\cite{mathlib2020}. Examples span mathematics and physics, from discrete geometry and number theory to nonlinear dynamics and chemical physics~\cite{hales2017kepler,anthropic2026fermat,immler2018lorenz,bobbin2024adsorption}. We bring this approach to physical design by formalizing published mechanisms together with the assumptions under which their predictions hold.

Category theory provides a language for composing physical models through their inputs and outputs~\cite{spivak2014,fong2019,censi2022}. In a string diagram, boxes represent laws and wires identify the quantities passed between them~\cite{selinger2010,coecke2017}. Related approaches have been used to compose models of circuits and reaction networks, describe material hierarchies, and translate biological mechanisms into engineered systems~\cite{baez2018,baez2017,spivak2011,giesa2012,giesa2015,brommer2016,marom2026}. A connected group of laws can be treated as a single component with its own inputs and outputs. For example, proofs that each law supplies valid inputs to the next can be combined to establish a domain guarantee for the assembly. A later model can use that guarantee directly once its own inputs are shown to satisfy the assembly's assumptions.

Here we introduce FormalGen, an autonomous multi-agent framework that brings this paradigm to scientific discovery and engineering design. Agents autoformalize published mechanisms by translating their governing laws, interfaces, and assumptions into mathematical models. They apply genetic operators to propose mutations that add, replace, or transfer laws. Lean~\cite{demoura2021} checks whether a claimed response follows from the resulting composition. The workflow advances accepted mutations to numerical testing and can use discrepancies to guide revision.

We evaluate FormalGen across 32 mechanisms. Ranking candidate connections by their physical interfaces and admissible domains recovers 100 connections with verified feasible states after examining 97\% fewer candidates on average than ranking by graph connectivity alone. We follow one accepted mutation through finite-element testing. The design brief asked how sunflower phototropism and the graded wall mechanics of bamboo could be combined in a stem whose rigidity changes as it bends toward light. Drawing on published models of sunflower bending~\cite{moulton2020} and bamboo stem rigidity~\cite{wegst2011}, the agents proposed that light-induced bending could reorient reinforcing fibers toward the stem axis. A search of the 2,225 papers in our corpora found no report of this coupling sequence. The composed model predicted a 2.46-fold increase in flexural rigidity with fiber alignment, and finite-element simulations gave a 2.45-fold increase between the two prescribed orientations.

The workflow also operated in the forward direction. For the other mutations, the agents began with existing mechanisms, proposed changes to their composition without a specified target response, and derived and checked the consequences of each change. Together, these results show how composition and formal verification extend AI-assisted, physics-grounded design. Agents can search for a coupling that produces a desired response or discover what a new composition predicts. In either direction, the prediction can be traced to the governing laws and checked under the assumptions that connect them. The proof-reuse benchmark illustrates how such work can accumulate. As the registry grew from 8 to 32 mechanisms, the number of individual law proofs grew fourfold, while the number of two-mechanism chains certified using those proofs grew 27-fold. This gives compositional design a basis for cumulative discovery. A mechanism derived in one investigation can become a building block for the next, which can use its verified properties to derive further testable predictions wherever its assumptions hold.

\section*{Results}
\label{sec:results}

\subsection*{Agentic Algorithm}
\label{sec:overview}

\begin{figure}[h]
    \centering
    \includegraphics[width=1.0\linewidth]{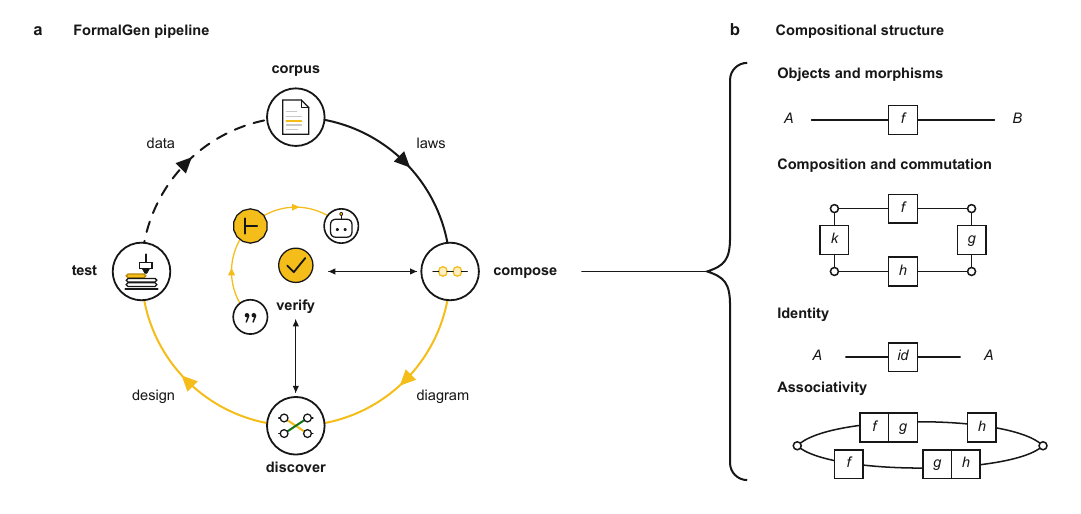}
    \caption{Overview of FormalGen.
\textbf{a}, The pipeline is a cycle of four stages around a verification stage.
The corpus of published papers supplies physical laws.
The compose stage writes each law as a morphism between typed objects and composes the morphisms into a string diagram.
The discover stage applies mutation operators to verified diagrams and returns candidate designs.
The test stage evaluates a design by finite-element simulation or by fabrication and measurement.
The dashed arrow marks the return of test data to the corpus.
Double arrows show that composed diagrams and candidate designs both pass through verification.
The three verification symbols denote source review (quotation mark), language-model generation and proof repair (agent) and proof checking by the Lean kernel ($\vdash$).
\textbf{b}, The compose stage uses the structure of a category.
A morphism $f$ maps an object $A$ to an object $B$.
Two morphisms compose when the codomain of the first is the domain of the second, and a diagram commutes when its two composites are equal, $g \circ f = h \circ k$.
The identity morphism $\mathrm{id}$ maps an object $A$ to itself and leaves every composite unchanged.
Composition is associative, $(h \circ g) \circ f = h \circ (g \circ f)$.
Boxes denote morphisms, wires denote objects, and open circles mark the objects shared by two composites.
    }
    \label{fig:pipeline}
\end{figure}

FormalGen takes a description of a mechanism or a desired response and works through four stages: corpus, compose, discover, and test (Fig.~\ref{fig:pipeline}a). The computational stages can run autonomously once the objective and test protocol are specified. A separate supervisory agent monitors the progress of each run and helps resolve problems as they arise. Researchers can intervene by supplying sources, revising the objective, proposing a missing coupling, or refining the test protocol.

In the corpus stage, an agent selects source papers from passages retrieved in two indexed collections. In this study, the bioinspired corpus contains 1,362 papers on biological and bioinspired structures, and the stimulus-responsive materials and mechanics corpus contains 863 papers. Researchers can also provide papers directly.

In the compose stage, a reconstruction agent records the governing laws, variables, assumptions, and evidence from the selected papers. A source-review agent checks whether the record covers the physics and scales required by the query. If a law is missing, the algorithm can search the corpus again. The mathematical reference specifies the allowed inputs and the response predicted by the source model. A composition agent assembles the laws into a candidate diagram that can be checked against this reference.

Figure~\ref{fig:pipeline}b shows how the composition agent assembles laws. Each wire represents an interface, or categorical object, and each box represents a law, or morphism. An interface specifies the quantities it carries, such as a scalar, a spatial field, or a signal that varies over time, and the values those quantities can take. Formally, an interface consists of typed quantities $X$ and a domain of allowed values $D_X\subseteq X$. A law $f\colon(X,D_X)\to(Y,D_Y)$ is defined for every input in $D_X$ and satisfies
\begin{equation}
f(D_X)\subseteq D_Y .
\label{eq:admissible}
\end{equation}
If the next law $g\colon(Y,D_Y)\to(Z,D_Z)$ satisfies $g(D_Y)\subseteq D_Z$, the two domain proofs give $(g\circ f)(D_X)\subseteq D_Z$. The first law's output guarantee supplies the condition needed to apply the second. When two paths between the same interfaces are intended to give the same result, as in the two routes in Fig.~\ref{fig:pipeline}b, commutativity expresses that claim as $g\circ f=h\circ k$.

Acceptance requires both source review and a certificate checked by Lean. When a proof fails, a repair agent uses Lean's diagnostics to revise it while keeping the definitions, assumptions, and theorem statement fixed. FormalGen stores accepted models with their diagrams, equations, assumptions, source evidence, and certificates.

In the discover stage, an agent modifies verified models using genetic operators such as addition, substitution, and crossover. It records which laws each candidate retains or changes. The same source review and Lean verification determine which candidates advance to testing.

In the test stage, FormalGen evaluates predictions of the accepted design using numerical simulation, implemented here with finite-element models. Simulation results can guide experimental validation. Results from either form of testing can be recorded in the corpus as evidence for future runs or returned to the compose stage to revise laws, parameters, or couplings.

\subsection*{Formalization and Verification}
\label{sec:formalization}

To compose physical mechanisms drawn from different papers, agents need a common language for specifying what each law requires and what it produces. A candidate connection must agree on the physical quantity being exchanged, its units, and whether it is a single value or varies in space or time. The output also has to satisfy the conditions under which the receiving law applies. A symbolic mathematical representation makes these requirements explicit and lets agents state compatibility and predictions as claims that Lean can check.

Category theory provides rules for compositional hierarchical design. Interfaces are represented as objects, and laws as morphisms between them. A composite has the same kind of input and output interface as an individual law, so a group of laws can serve as a component in a larger model. The interfaces also show where a new coupling law is needed to supply a quantity the receiving law requires. Lean checks whether the output of each law lies in the next law's admissible domain (Eq.~\eqref{eq:admissible}) and whether the claimed response follows from the assembled laws under the stated assumptions.

\begin{figure}[h]
    \centering
    \includegraphics[width=1.0\linewidth]{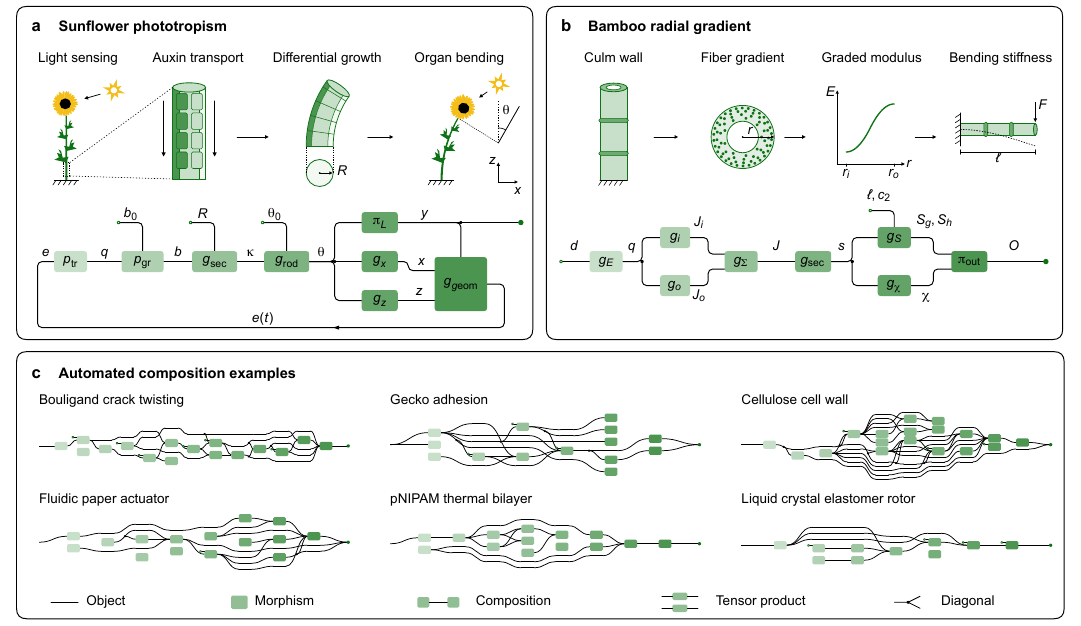}
    \caption{
    Verified formalizations of published mechanisms.
\textbf{a}, Schematics (top) and string diagram (bottom) of sunflower phototropism~\cite{moulton2020}.
Transport $p_{\mathrm{tr}}$ maps the light-direction signal $e$ to the auxin asymmetry coefficient $q$.
The growth, section, and rod maps $p_{\mathrm{gr}}$, $g_{\mathrm{sec}}$, and $g_{\mathrm{rod}}$ yield the growth coefficient $b$, intrinsic curvature $\kappa$, and tangent angle $\theta$, given initial value $b_0$, stem radius $R$, and base angle $\theta_0$.
The projection $\pi_L$ returns tip inclination $y$, $g_x$ and $g_z$ give centerline coordinates $x$ and $z$, and $g_{\mathrm{geom}}$ updates the feedback signal $e(t)$.
\textbf{b}, Schematics and string diagram of bamboo's radial modulus gradient~\cite{wegst2011}.
The profile map $g_E$ maps the wall radii and endpoint Young's moduli in $d$ to the coefficients $q$ of $E(r)$ on $[r_i,r_o]$.
The integrals $g_i$ and $g_o$ yield modulus-weighted radial moments $J_i$ and $J_o$, which $g_{\Sigma}$ sums to $J$.
The section map $g_{\mathrm{sec}}$ returns the pair $s$ of graded and homogeneous sections with equal mean modulus.
The bending test $g_S$ gives stiffnesses $S_g$ and $S_h$ for length $\ell$, end-condition coefficient $c_2$, and load $F$.
The map $g_{\chi}$ gives the rigidity ratio $\chi$ (gradient factor), and $\pi_{\mathrm{out}}$ returns $O$.
\textbf{c}, Six mechanisms use sequential composition, parallel composition (tensor product), and shared-input copying (diagonal).
Boxes denote morphisms or relations, and wires denote interface objects.
Open ports mark parameters, filled circles mark readouts, and green shading shows composition order.
    }
    \label{fig:mechanisms}
\end{figure}

The agents formalized and verified 32 mechanisms. Sunflower phototropism and bamboo mechanics illustrate how the same representation captures distinct dependencies across scales (Fig.~\ref{fig:mechanisms}a,b). Both models are reused in the combined design that follows.

The agents formalized a sunflower phototropism mechanism described in the literature~\cite{moulton2020} as the string diagram in Fig.~\ref{fig:mechanisms}a. Growth accumulates the effects of auxin asymmetry over time, so the diagram passes the time-varying signal at each position to the growth law. The tip position and orientation feed back into light sensing. A dedicated Lean proof establishes that the composed model and the encoded source equations impose the same conditions on trajectories under the stated assumptions.

The agents formalized a separate model of bamboo mechanics~\cite{wegst2011} as the string diagram in Fig.~\ref{fig:mechanisms}b. The diagram supplies the same radial modulus profile to the inner- and outer-wall calculations and combines their contributions to bending rigidity. Lean checks that the intermediate quantities meet the required conditions, including positive moduli, and that the assembled laws yield the stated bending stiffnesses and the rigidity ratio relative to a uniform wall of the same mean modulus. Supplementary Methods, section~1, gives the governing equations and assumptions for both models.

Figure~\ref{fig:mechanisms}c shows six further multiscale or hierarchical systems, with three examples from each corpus. These include Bouligand crack twisting, gecko adhesion, cellulose cell wall mechanics, a fluidic paper actuator, a thermally responsive hydrogel bilayer, and a liquid crystal elastomer (LCE) rotor. These examples expose the intermediate relations through which local properties determine a larger-scale response.

These representations follow established diagrammatic conventions for sequential and parallel composition and explicit copying of shared inputs~\cite{selinger2010,fong2019}.
Sequential connections expose the dependence of one calculation on another, branches identify quantities reused by several components, and converging paths in these models identify responses entering a common balance or geometric relation.
Parallel subcompositions can retain shared parameters and joint constraints, which must be preserved when the model is modified.
For hierarchical design, a connected subcomposition can be treated through its external interfaces~\cite{vagner2015}, while retaining access to its internal laws and assumptions.
Multiscale reuse therefore depends on retaining the relations that connect the constituent response to the observable at the larger scale.

The registry makes a verified constitutive law or a connected group of relations available as a design building block with explicit conditions for reuse.
Replacing such a block requires compatibility with the receiving interfaces and preservation or revision of the affected coupling relations.
Changes to the assembled response require renewed verification of the corresponding claims.

\subsection*{Mutations and Discovery}
\label{sec:discovery}

We use verified compositional building blocks to formulate material design hypotheses through operations inspired by genetic algorithms~\cite{holland1975}. These operations, termed mutations, edit the laws or connections in a diagram. The retained laws, admissible interfaces, and source evidence constrain how the edited model can be assembled. A reused proof applies wherever its assumptions still hold. The new connections and the claimed response of the assembled model require verification.

\begin{figure}[h]
    \centering
   \includegraphics[width=1.0\linewidth]{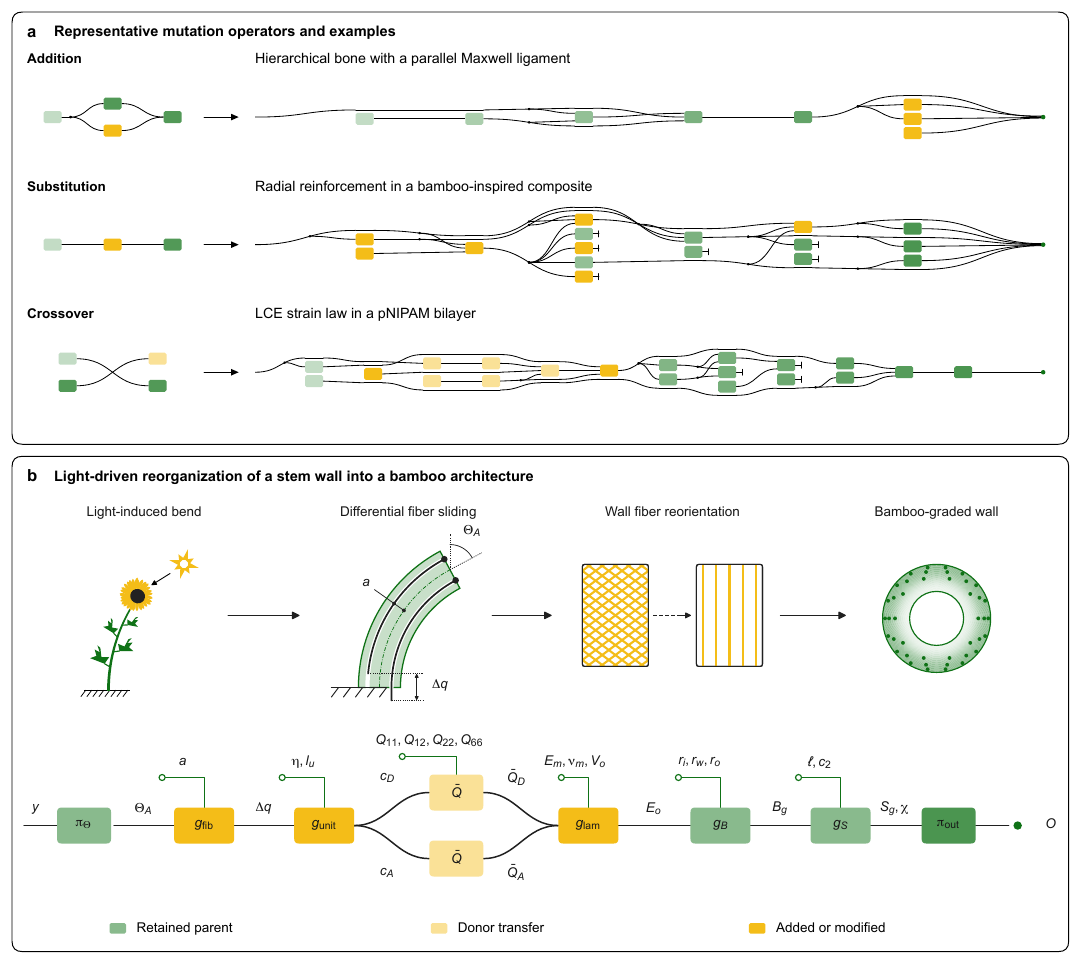}
    \caption{
Mutation of verified compositions.
\textbf{a}, Agent-generated examples of individual mutation operators. Each row shows the operator (left) and its verified composition (right).
Addition extends the hierarchical bone composition with a parallel Maxwell ligament and a force-summing assembly map.
Substitution replaces the radial reinforcement map of the bamboo-inspired composite at fixed domain and codomain.
Crossover transfers the constitutive subdiagram of a liquid crystal elastomer (LCE) into a poly(\textit{N}-isopropylacrylamide) (pNIPAM) bilayer through entry and exit maps.
\textbf{b}, A user-guided design combines sunflower phototropism~\cite{moulton2020} with the bamboo radial gradient~\cite{wegst2011}. The schematics (top) and string diagram (bottom) show how light-driven bending reorients the wall fibers and increases the predicted stem rigidity.
The maps convert the bend angle to fiber sliding and rotation ($\pi_{\Theta}$, $g_{\mathrm{fib}}$, $g_{\mathrm{unit}}$), fiber orientation to wall modulus ($\bar Q$, $g_{\mathrm{lam}}$), and wall modulus to stem rigidity and bending stiffness ($g_B$, $g_S$). Supplementary Methods, section~2, defines the map parameters.
Symbols follow Fig.~2, and terminal bars discard a wire.
    }
    \label{fig:mutations}
\end{figure}

Figure~\ref{fig:mutations}a shows three mutations generated by agents in automated runs, each illustrating a single operator. Addition extends a model with a component and the maps that couple it to the retained laws. In the bone-inspired scaffold~\cite{zhang2014damping}, a Maxwell ligament is added in parallel. The two branches share the imposed strain, and an assembly map sums their axial forces. The added branch increases dissipation per cycle at fixed strain and frequency, with additional material.

Substitution replaces a law or group of laws within an existing input--output boundary. In the bamboo-inspired composite~\cite{wegst2011,sato2017}, the agents substitute a radial profile that moves reinforcement outward while holding total fiber volume fixed. The replacement supplies the same type of profile to the retained section calculation, which yields greater flexural rigidity under the stated assumptions.

Crossover transfers a connected subdiagram from one model to another. Entry and exit maps connect the transferred laws to the recipient's interfaces. In the poly(\textit{N}-isopropylacrylamide) (pNIPAM) bilayer, these maps express a liquid crystal elastomer (LCE) strain law relative to the bilayer's reference state while retaining its section mechanics~\cite{yarali2025,zhang2025rotor}. The transferred law reverses the predicted bending direction at the specified temperature. Lean verifies all three response changes under their respective assumptions.

Figure~\ref{fig:mutations}b shows a design generated through several compositional edits in response to a user query. The user asked FormalGen to couple sunflower phototropism with the graded-wall mechanics of bamboo. The sunflower model supplies a bend angle, and the bamboo model requires a radial modulus profile. The researcher set aside one verified candidate because it left the bamboo wall unchanged. A later candidate passed Lean and failed the simulation test, with rigidity at $53^\circ$ above the prediction by 53.8\% and 12.3\% at two design points. The agents traced the error to the wall law, which treated each fiber layer as free, and replaced it with the law of a bonded laminate from the corpus. The revised composition was accepted on the fourth pass and passed the same test (Supplementary Methods, section~2).

The accepted composition retains the bend-angle output of the sunflower model and the section and beam calculations of the bamboo model. Between them, the agents assembled a chain of laws in which bending pulls two sliding fibers by different amounts. A transmission uses the difference in travel to turn the reinforcing fibers toward the stem axis. A bonded-laminate law converts the new fiber orientation into the wall modulus profile used by the bamboo model. Under the stated assumptions, including matched Poisson coupling, Lean checks that the higher axial modulus of the aligned fiber layer produces greater flexural rigidity and bending stiffness. For the reference design, rotating the fibers from $53^\circ$ to $0^\circ$ raises the predicted rigidity from 6,049 to 14,882\,N\,m$^2$, a factor of 2.46 (Supplementary Methods, section~3).

The resulting design predicts that light-driven bending can stiffen the same stem by reorienting its wall fibers. A text screen of the 2,225 papers in the two corpora found no report of this coupling sequence, supporting the novelty of the proposed mechanism (Supplementary Methods, section~2). By composing the verified models with the intervening laws, the agents derived a testable prediction and identified conditions for realizing it, including the work and clearance needed to turn the fibers.

\subsection*{Testing Predictions from Accepted Designs}
\label{sec:simulation}

\begin{figure}[t!]
    \centering
    \includegraphics[width=1.0\linewidth]{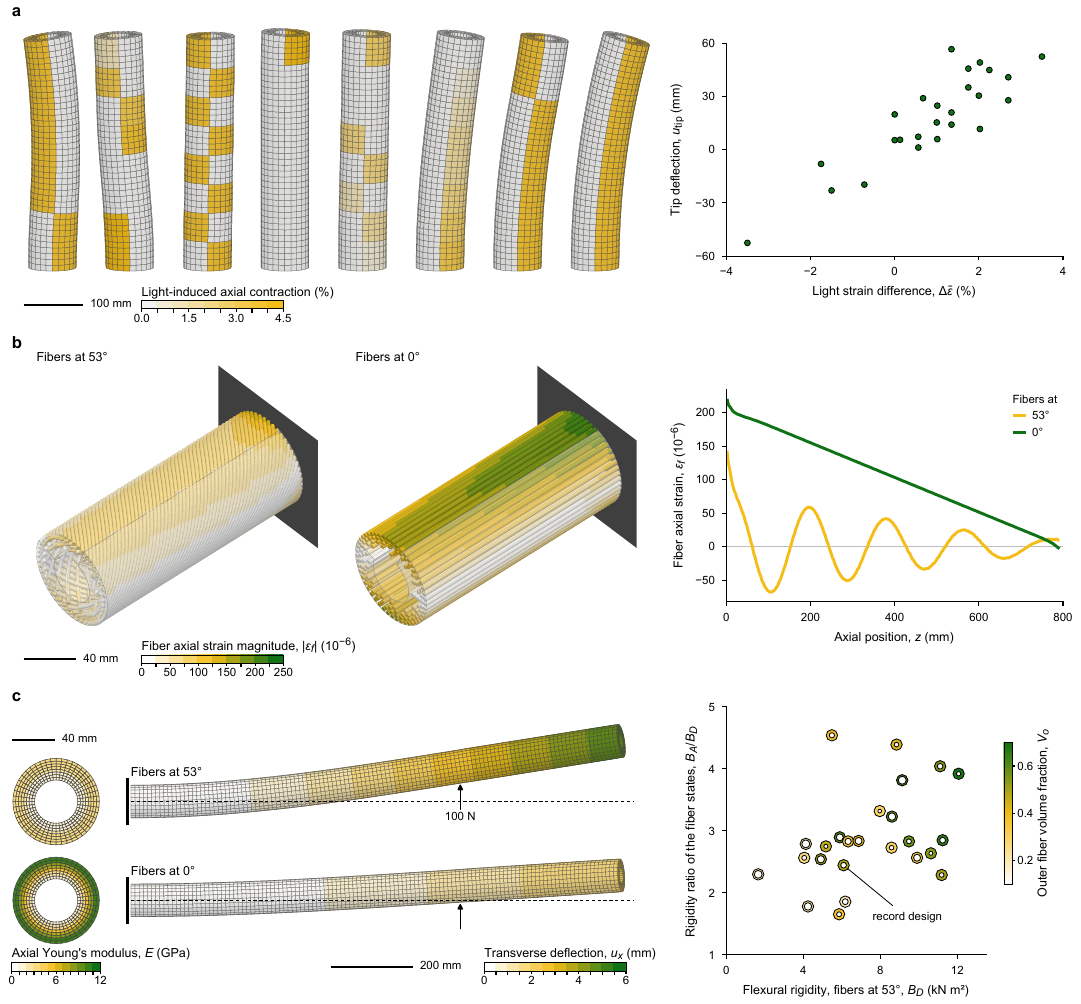}
    \caption{Finite-element simulations of the light-driven stem with reorientable wall fibers.
\textbf{a}, Light-responsive segment under eight illumination patterns, coloured by
light-induced axial contraction (left), and tip deflection $u_\mathrm{tip}$ against the
mean light strain difference of the two sides, $\Delta\bar{\varepsilon}$, for 25 patterns
and light levels (right).
\textbf{b}, Tube wall with its fibers resolved as 137 bundles at $53^{\circ}$ and
$0^{\circ}$ under a 100\,N tip load, coloured by axial strain magnitude (left), and axial
strain $\varepsilon_f$ along one outer bundle in each state (right).
\textbf{c}, Whole member under a 100\,N transverse load with fibers at $53^{\circ}$ and
$0^{\circ}$, coloured by transverse deflection $u_x$, with wall
sections coloured by axial Young's modulus $E$ (left), and rigidity ratio of the two fiber
states, $B_A/B_D$, against $B_D$ for 24 designs and the record design, with each marker a
wall section coloured by outer fiber volume fraction $V_o$ (right).}
    \label{fig5}
\end{figure}

FormalGen passes an accepted formalized mutated composition to the next stage for in silico testing. For the stem design, the agent ran finite-element simulations to test whether light-induced bending produces the predicted fiber travel, how fiber alignment changes load transfer in the wall, and how much the stem stiffens between the two prescribed orientations (Fig.~\ref{fig5}).

Tip deflection depends on where the light falls (Fig.~\ref{fig5}a).
Across 25 illumination patterns and light levels, patterns with the same mean strain difference between the two sides produce different tip deflections.
At a difference of about $2.0\%$, simulated tip deflections are 11.7\,mm with the light near the tip, 30.5\,mm with uniform light on one side, and 49.2\,mm with the light near the base.
The sliding law $g_{\mathrm{fib}}$ uses rotation at the anchors, so it can be tested across illumination patterns even as the overall deflection changes.
An initial test using a plane fitted to the tip section missed the acceptance criteria. In 12 held-out cases using rotation at the fiber anchors, the simulated sliding displacement agreed with $2a\Theta_A$ within $1.0\%$ in ten cases and within $0.01$\,mm in the two cases with near-zero rotation.

Resolving the individual fiber bundles shows why alignment changes load transfer (Fig.~\ref{fig5}b).
The tube with explicit bundles and the continuum wall agree in tip stiffness to within $0.2\%$ at both fiber angles.
The axial strain in the plotted straight bundle falls approximately linearly along the tube. In the helical bundle, strain alternates in sign as it crosses between tension and compression. The largest bundle strain magnitude at the clamp was $221\times10^{-6}$ for aligned fibers and $149\times10^{-6}$ for helical fibers.

The agent compared the predicted rigidities with the continuum simulation results and tested how the change in wall properties affects deflection of the complete member (Fig.~\ref{fig5}c).
For the reference design, the simulated flexural rigidity is 6,090\,N\,m$^2$ with fibers at $53^\circ$ and 14,903\,N\,m$^2$ at $0^\circ$, within $0.7\%$ and $0.2\%$ of the model predictions, and their ratio is 2.447 against 2.460.
Under a 100\,N transverse load applied at the junction of the graded base and the light-responsive segment, the simulated deflection at that junction was 2.85\,mm at $53^\circ$ and 1.18\,mm at $0^\circ$, a stiffness ratio of 2.41. The whole-member values are from a 11,520-element mesh (Supplementary Methods, section~3).
Across 24 designs sampled over the declared domain, the simulated rigidity is within $1.8\%$ of the composition in both states, while the ratio ranges from 1.7 to 4.5.

For the longer cantilevers, the bamboo-inspired gradient increased stiffness relative to a uniform wall with the same mean modulus.
At $0^\circ$, the graded tube was 1.143 times as stiff as the uniform tube at 800\,mm and 1.157 times at twice that length, compared with a predicted factor of 1.167.
The closer agreement for the longer tube is consistent with a smaller contribution from shear deformation. At a length of five outer radii, the stiffness ratio between the two fiber orientations fell to 1.95, and the aligned graded tube was slightly less stiff than its uniform comparator.
The stiffness simulations prescribe the fiber angle of each state. In a separate finite-strain test of a bonded strip, the twelve fiber units ended at angles of $16^\circ$ to $18^\circ$ with $0^\circ$ as the target, and the matrix exceeded the specified strength limit. These results motivate a wall that releases the fibers during rotation. With the fibers prescribed at $16.2^\circ$ and $18.1^\circ$ in an intact wall, the simulated flexural rigidity of the reference design is 2.18 and 2.12 times its value at $53^\circ$, or 89\% and 87\% of the aligned value (Supplementary Fig.~1). Together, the simulations support the predicted stiffening between prescribed orientations and guide experimental design. An experiment can measure how far the fibers rotate under illumination and how much that rotation changes the flexural rigidity of the stem.

\subsection*{Benchmarking compositional graph representations}
\label{sec:connections}

How many connections suggested by a knowledge graph can actually supply the inputs of another physical law? We compared two representations of the same 32 mechanisms and 295 objects. The baseline knowledge graph proposes connections when two quantities share a name or have a recorded relationship. The compositional graph uses the underlying model records to ask whether an output can supply a specific input and whether the joined models can share a state. Using the same mechanisms in both graphs lets us measure what the composition rules add to the search.

\begin{figure}[t!]
    \centering
    \includegraphics[width=1.0\linewidth]{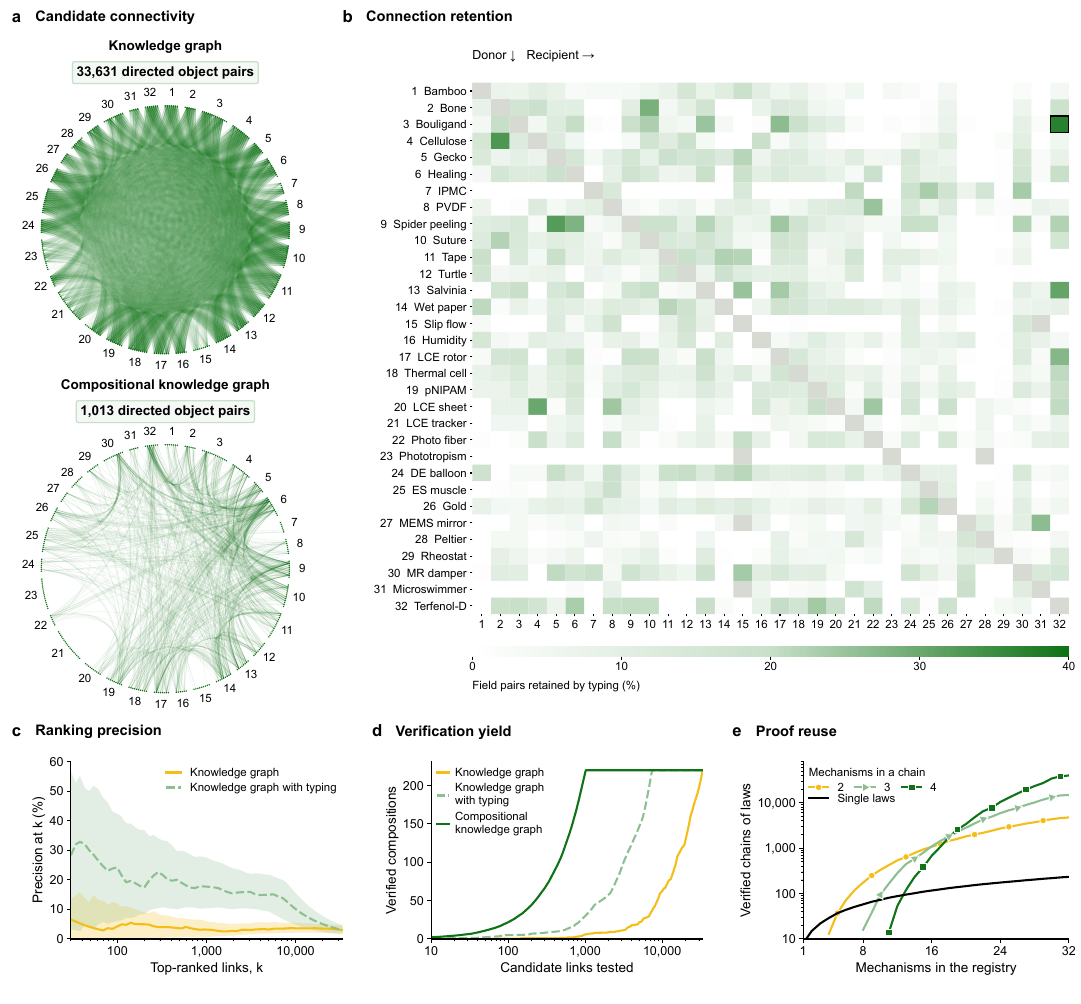}
    \caption{
Compositional knowledge graph benchmarking.
\textbf{a}, Directed object pairs linked in the knowledge graph (top) and in the compositional knowledge graph (bottom) for 32 source compositions with 295 objects.
Dots are objects, and numbers are the mechanisms in \textbf{b}.
The knowledge graph links an output field to an input field whose quantities are the same or related.
The compositional knowledge graph keeps object pairs that supply every recipient input with a matching quantity, SI dimension, and mathematical form and pass the joint linear feasibility screen.
\textbf{b}, Percentage of knowledge-graph field pairs that pass typing, for each mechanism pair (donor in row, recipient in column).
Typing retains 36,120 of 418,793 field pairs.
Gray cells are self-pairs or pairs without a link.
\textbf{c}, Precision at $k$, the percentage of the $k$ highest-scoring knowledge-graph object pairs that belong to the compositional knowledge graph.
An object pair is scored by its number of linked field pairs (knowledge graph) or of typed field pairs (knowledge graph with typing).
The base rate is 3.0\% (1,013 of 33,631).
The knowledge-graph score stays at the base rate, and the typed score gives about 20\%.
Shaded bands are 95\% bootstrap intervals.
\textbf{d}, Cumulative count of direct connections with a verified feasible state against candidate rank, averaged over random orderings of ties. Each ranking is scored against the states verified among pairs that passed the direct-connection screen. With the compositional knowledge graph first, the average count reaches 100 at rank 455, compared with rank 15,166 by knowledge-graph score.
\textbf{e}, Proof reuse as the registry grows. Curves show counts of individual laws with verified domain-preservation proofs and of chains spanning two, three or four distinct mechanisms. Each counted chain has a Lean-checked proof assembled from its constituent law proofs, a stated output domain and at least one input satisfying the domain conditions throughout the chain. Counts are medians over 1,000 random orderings of the 32 mechanisms.
    }
    \label{fig:connectivity}
\end{figure}

The first check asks whether quantities can be passed directly from one law to the next (Fig.~\ref{fig:connectivity}a,b). The output must represent the same physical quantity, have the same SI dimensions, and have the mathematical form the next law expects. A single temperature in kelvin, for example, cannot by itself supply a temperature as a function of time. We refer to this comparison as interface typing. It retains 7,379 of 33,631 directed object pairs. These pairs have a starting point for direct transfer, since the donor can supply at least one of the recipient's fields.

At the level of individual fields, 36,120 of 418,793 proposals pass this check, a $91.4\%$ reduction. Much of this reduction comes from proposals involving related quantities that would require a law to convert one into the other. Among the 41,885 proposals that already share a quantity label, checking dimensions and mathematical form removes $13.8\%$. Direction matters because an output from one model can supply an input to the other even when the reverse transfer is impossible. The knowledge graph proposes 1,144 field connections from sunflower to other mechanisms, including 44 to bamboo. None can be passed directly. The sunflower and bamboo mechanisms therefore need an explicit coupling law, as the design example illustrates.

Of the 7,379 pairs with a matching field, 6,146 leave at least one required input unfilled. The next check asks whether the values required by the remaining 1,233 pairs can occur together. Of these pairs, 136 fail bounds on individual quantities and 38 lack a recorded domain. Joint linear constraints exclude 46 more pairs that pass the separate bounds. The remaining 1,013 pairs, $3.0\%$ of the original proposals, form the compositional knowledge graph. The screen can omit nonlinear conditions, which are checked in the subsequent state search. The 46 additional exclusions show why compatible ranges must be checked together: the values may fit each range separately and still be impossible to realize in the same state.

The number of knowledge-graph links between two objects gives little indication of whether they pass these checks (Fig.~\ref{fig:connectivity}c). Of the top 100 and 1,000 pairs ranked by that score, $5.0\%$ ($95\%$ interval 0 to $11\%$) and $3.2\%$ (0.7 to $5.7\%$) pass, respectively, against a base rate of $3.0\%$. Ranking by matching fields after typing raises those percentages to $22\%$ (6 to $42\%$) and $18\%$ (11 to $27\%$). Typed matches therefore give a more informative ranking for the feasibility screen.

We searched the 1,013 pairs that passed the direct-connection screen and found feasible states for 220. The Lean kernel checked 203 states, and interval arithmetic verified the other 17. To compare the rankings, we measured how many candidates each had to examine to recover 100 of the 220 verified direct connections. The average count reached 100 at rank 455 when pairs in the compositional graph came first, compared with rank 3,148 for typed matches and rank 15,166 for knowledge-graph connectivity alone (Fig.~\ref{fig:connectivity}d). Couplings that require an additional input or a new law are separate design candidates whose interfaces and physical conditions must be checked for the resulting composition.

We introduced faults into models from 5 mutations across 4 parent mechanisms. These diagnostics used sampled numerical evaluations alongside the separate Lean checks of the encoded models.
The connectivity screen missed all 30 faulty variants. Dimension checks caught 5, and checks of the retained laws and their domains caught 28.
All screens accepted the original models and their algebraically equivalent rewrites.
The remaining defects failed subsequent physical checks. The 4 controls with the added feature disabled passed those checks and failed on the design objective.
Checking the retained laws catches faults that survive dimensional checks. The disabled-feature controls show why the requested response also needs its own check.

We also tested whether the proof for a complete path can be assembled directly from proofs for its constituent maps (Supplementary Methods, section~4). Across 31 certificates, Lean accepted the assembled domain-preservation proof for all 794 chains of at least two maps without proof search. When given the same end-to-end statements without the local proofs, the default tactics proved 268 under the specified search budget. The existing map proofs came from the full proof-repair workflow. The result demonstrates proof reuse: once individual maps have been verified, their domain guarantees can be carried through a larger calculation mechanically.

When a new mechanism connects to one already in the registry, proofs from both mechanisms can be combined to certify the resulting chain of laws. We measured how the number of chains that could be certified in this way grew with the registry (Fig.~\ref{fig:connectivity}e). Registries of eight mechanisms held a median of 58 proofs for individual laws and supported a median of 175 chains spanning two mechanisms. With all 32 mechanisms, the number of individual proofs grew roughly fourfold to 234, while the number of two-mechanism chains grew 27-fold to 4,766. The same registry supported a further 14,872 chains spanning three mechanisms and 40,015 spanning four. For all 59,653 chains, we assembled the domain-preservation proof from existing proofs without further proof search. A law proved once can therefore carry its guarantee into an expanding range of designs, provided the conditions at each connection are met.

The graph serialization control reconstructs the model from graph records and reproduces the connection and mutation checks described in Supplementary Methods, section~4. A graph that encodes the same maps and interfaces can support the same operations. The operational advantage comes from equipping a connection with a rule for composing both the physical maps and their proofs. This lets a verified chain serve as a unit of construction in a new design.

\section*{Discussion}
\label{sec:discussion}

FormalGen demonstrates a transition from open-ended reasoning to formalism in AI-driven design, in which AI-generated material design hypotheses are made composable and formally verifiable. Agents construct hypotheses from literature-derived mechanisms, modify their component laws and connections, and submit claims about the resulting behavior to Lean. A verified design therefore carries an explicit mathematical account of how its response follows from its governing laws and assumptions. This connects generative hypothesis formation to physical reasoning that can be inspected, modified, and checked. Verification becomes part of constructing a design hypothesis and deciding which proposed changes are mathematically justified.

Governing equations become reusable design components when their inputs, admissible domains, and coupling conditions are specified. FormalGen organizes this information around objects, morphisms, and their compositions, giving explicit meaning to both a proposed connection and a change to the model. The representation determines what must remain consistent when components are added or replaced and which response claims require renewed verification. In the benchmarks, checking interfaces and joint conditions brought feasible direct connections earlier in the search. Combining the proofs for individual laws established the domain guarantees of larger assemblies. Together, these results show how a formal representation helps agents find candidate connections and build on verified work.

This representation opens a design space that includes both component behavior and mechanism architecture. A design can change the distribution of material properties, introduce a new interaction, or transfer a sequence of physical processes between systems. The coupled sunflower--bamboo model illustrates how an engineered material could stiffen as it bends toward light. Reorienting the wall fibers connects the light-driven bend to the change in rigidity. Such a hypothesis can be assessed through the laws that connect these processes and the response they produce together. A verified subcomposition can subsequently serve as a composite morphism in another model, provided its assumptions remain satisfied. This creates a basis for cumulative design in which an established mechanism becomes a building block for further hypotheses.

The simulations support the predicted stiffening between prescribed fiber orientations. A physical realization must establish whether light-driven bending can rotate the fibers far enough and whether the wall can hold them in place. The composed model makes these requirements explicit, giving experiments specific conditions and responses to measure.

The broader opportunity is an AI-driven design process that accumulates reusable physical knowledge as it explores. An expanding registry could allow agents to search across mechanisms from different sources, construct candidate architectures toward a specified response, and use verified claims or violated constraints to guide subsequent mutations. A next computational step is to measure how this mathematical structure and verification feedback affect proposal quality across independently assessed design tasks. A corresponding physical step is to test selected hypotheses and attach the measured responses and operating conditions to their model records. Each evaluated design could then expand the mechanisms and evidence available to later searches, supporting physics-driven hypothesis generation that becomes more informed through continued use.

Although we demonstrate FormalGen in materials design, the approach could extend to other fields in which mechanisms can be connected through shared quantities and explicit assumptions. In a chemical reaction network, for example, a relation derived for one set of coupled reactions could serve as a component in a larger network once the conditions for its use are established. A researcher could then use that relation to derive and test a prediction about the larger system. This is the broader prospect for cumulative discovery. A mechanism developed to answer one scientific question can become a building block for the next, carrying with it a precise account of what has been proved and under which conditions. Formal verification lets each connection, once proposed, be checked, tested, and built upon, extending to physical design the transition now under way in mathematics.

\section*{Methods}
\label{sec:methods}

\subsection*{Corpus acquisition}

We assembled two literature corpora from the Web of Science Core Collection.
The bioinspired search used the All Fields query \texttt{biological material mechanical hierarchical structure}, restricted to English and records indexed through 23 March 2026.
The retained bioinspired collection contained 1,362 full-text papers.
The stimulus-responsive materials and mechanics search combined terms for material classes, physical mechanisms, mathematical models, and experimental evidence.
The searches for the stimulus-responsive materials and mechanics corpus covered all publication years and were run on 11 August 2026.
The retained collection contained 863 full-text papers.
The full search queries are provided with the code.

Agents received the full PDFs directly.
Text extracted with \texttt{pdftotext -layout} served as a page-labeled quotation index for checking cited evidence against the sources.

\subsection*{Models and software libraries}

We used \texttt{gpt-6-astra} for reconstruction, composition, source review, proof repair and supervision of each run. Lean checked each candidate certificate. When a check failed, the proof-repair agent used the compiler diagnostics to revise the proof. Certificate acceptance required a successful Lean check. The supervisory agent restarted runs that stopped, repaired defects in the workflow code and prompts, and corrected a model record when a loop stopped without an accepted result. Of the 21 runs with complete records, 8 needed no change to code, record, or proof.

Custom Python code coordinated model calls through the OpenAI Python SDK and Responses API. The notebooks specified Python 3.11. Recorded calls within the loops used maximum reasoning effort, JSON-object output and a limit of 128,000 output tokens per call. Lean 4.12.0 and mathlib provided the proof environment~\cite{demoura2021,mathlib2020}. Prompts, responses, settings and compiler diagnostics were saved as JSON. SHA-256 hashes linked accepted model records to their source documents and certificates.

\subsection*{PDF-to-composition algorithm}

From the supplied sources and goal, the workflow automatically generated a physical reference recording the supported equations and assumptions. It then generated a mathematical reference with variables, shared parameters, allowed inputs, and formulas for the reference response.

The composition agent received the PDFs, physical reference, and the input and output specifications from the mathematical reference. The direct reference formulas were withheld. It proposed the individual laws, their order of evaluation, and which paths should agree. A source-review agent assessed the resulting model before seeing the Lean verdict. Acceptance required both source reviews and a checked certificate. Failed reviews prompted revision of the model record, and failed proof attempts could be sent for proof repair. Acceptance came at the first composition attempt for 11 of the 21 mechanisms with complete run records, at the second for 8, and at the third for 2, and the researchers edited no model record, proof, or code.

The compiler assembled each proposed path by matching successive source and target objects.
\begin{lstlisting}[
    basicstyle=\small\ttfamily,
    breaklines=true,
    columns=fullflexible,
    keepspaces=true,
    showstringspaces=false
]
def compose(path, morphisms, source_object):
    expression, current_object = "x", source_object
    for name in path:
        morphism = morphisms[name]
        assert morphism["src"] == current_object
        expression = apply(name, expression)
        current_object = morphism["dst"]
    return expression, current_object
\end{lstlisting}
This pseudocode uses \texttt{x} for the input and \texttt{apply} for a symbolic function call with shared parameters.

\subsection*{Formal verification}

For supported algebraic and finite-dimensional dynamical models, the compiler generated Lean definitions and theorem statements from the model records (Supplementary Methods, section~5).
For each candidate morphism, every input allowed by the source object had to produce an output allowed by the target object.
The claims also checked physical dimensions and restrictions needed to evaluate the expressions, such as nonzero denominators.
These restrictions were extracted before simplifying the expressions.

Lean also checked that the composed response matched the mathematical reference and that alternative paths declared equal gave the same outputs.
These claims applied to all inputs and parameters within the stated conditions.
Each object also required at least one allowed state and parameter choice.
For finite-dimensional dynamical models, further proofs checked that morphisms transformed differentiable solutions of the source model into solutions of the target model.

Fixed definitions and proof templates encoded composition, identity, and associativity.
When the templates did not resolve a proof, an AI agent used Lean's error messages to repair it while the generated definitions, assumptions, and theorem statements remained fixed.
A certificate was accepted only after successful Lean compilation and an audit excluding unfinished proofs and additional axioms.
The registry stored each verified composition with its interfaces, equations, source evidence, assumptions, and Lean certificate.

\subsection*{Mutation}

A mutation request described the parent composition types and the response to establish. Supplementary Methods, section~5, gives the mathematical forms of addition, substitution, and crossover and the conditions checked at their interfaces.
The workflow passed each candidate through the same reference-generation, source-review, composition, and Lean stages used for the source mechanisms.
Its record labels every law as printed in a source, derived, standard, or assumed, and keeps the parent maps that are retained without change.
In the light-driven stem design, the researcher set aside one candidate after it passed source review and Lean. The agents screened the remaining candidates through source review, proof checking, and simulation as applicable. One candidate that passed Lean failed the simulation test, and the agents replaced its wall law before the accepted pass.
Supplementary Methods, section~2, lists the passes and their outcomes.

\subsection*{Graph construction and comparison}

We compared candidate connections among the same 32 mechanisms and 295 objects in both graph representations.
The knowledge graph linked objects to their fields and quantity labels, with directed relations extracted from equation dependencies and declared physical relations.
The compositional graph retained the corresponding equations, parameters, domain conditions, and proof references.

We standardized capitalization and punctuation in quantity labels and applied a fixed alias rule.
Each field retained its original label, units, and mathematical type.
The knowledge graph proposed connections between mechanisms when the output and input labels matched or were linked by one recorded directed relation.
For each field $x$, we defined the comparison signature
\[
s(x)=\Bigl(
\underbrace{N(\ell_x)}_{\text{quantity label}},\;
\underbrace{\mathbf d_x}_{\text{SI dimensions}},\;
\underbrace{c_x}_{\text{mathematical type}}
\Bigr).
\]
Here $N$ standardizes the recorded label $\ell_x$, $\mathbf d_x$ contains the powers of the SI base units, and $c_x$ is the declared mathematical type.
Direct-transfer candidates required matching signatures, $s(x)=s(y)$.
An object pair was retained when at least one field pair passed the rule.

When an output object could supply every required input and both objects had recorded domains, Z3 5.1.0 checked whether one assignment to the shared quantities could satisfy their linear constraints together.
Omitted nonlinear constraints were recorded.
The pairs that passed formed the compositional knowledge graph.

The two baseline rankings scored each object pair by the number of linked field pairs in the knowledge graph or the number of field pairs that passed typing.
For these rankings, precision at $k$ was the share of the top $k$ pairs that belonged to the compositional knowledge graph. Precision was averaged over 20 random tie orderings. The 95\% intervals were calculated from 400 bootstrap resamples of the 32 mechanisms, with each pair weighted by the product of the resampling counts of its donor and recipient mechanisms.

For the recovery benchmark, a third ranking placed pairs that passed the joint screen first, followed by other pairs with a typed match for every required recipient input, then all remaining pairs. Within each group, pairs were ranked by the fraction of required recipient inputs with a typed match.
A search over the 1,013 pairs that passed the joint screen found 220 feasible states, using exact rational arithmetic and rigorous intervals for transcendental functions. Lean checked the bounds and conditions at 203 states, and interval arithmetic verified the other 17.
All three rankings were scored against the same 220 verified connections. At each rank, the cumulative count was averaged over 50 random orderings of ties. The reported rank is the first at which that mean reached 100.

We used these 220 verified connections to enumerate chains of laws spanning two, three or four distinct mechanisms (Fig.~\ref{fig:connectivity}e). Each transfer bound donor output fields to the inputs read by the recipient. Within each intermediate mechanism, a sequence of recorded laws had to lead from the object that received one transfer to the object that supplied the next. We combined the domain-preservation proofs of the constituent laws in composition order, without further proof search. Lean checked that the final state lay in its stated domain whenever the parameters and source state were admissible and each transferred state satisfied the receiving object's domain conditions. We counted a chain when Lean accepted this proof, the final object had a stated domain condition, and we found at least one state satisfying the hypotheses throughout the chain. To measure how proof reuse varied with registry size, we sampled 1,000 random orderings of the 32 mechanisms. For each registry size $n$, we report the median counts of individual law proofs and qualifying chains among the first $n$ mechanisms.

\subsection*{Finite-element simulation}

The agent used CalculiX 2.23~\cite{dhondt2004} for the Fig.~\ref{fig5} simulations, with reduced-integration 20-node hexahedra for the solid meshes. A separate finite-strain model implemented in JAX tested whether fibers bonded to the matrix could rotate into alignment (Supplementary Methods, section~3). The agent used scripts to read stiffness predictions from the accepted model record, generate finite-element inputs, and compare the simulation results with the predictions. The agent specified the readouts and acceptance criteria before the corresponding solves or, for analyses of existing solutions, before extracting the readouts. Supplementary Methods, section~3, reports the initial failures and the results of the follow-up tests.

The wall model used a tube with radii of 20, 30, and 40\,mm and four, six, or eight radial layers of alternating fiber angle. The material tensor of each layer combined the rotated lamina and matrix stiffnesses at the recorded fiber fraction at the layer midpoint. Fiber orientations of $53^\circ$ and $0^\circ$ were prescribed. Flexural rigidity was calculated as $B=ML/\phi$, using the moment $M$ at an end rotation $\phi=0.01$\,rad for a tube of length $L=400$\,mm. The mesh study used three resolutions up to 30,720 elements. To test robustness across the declared domain, the agent generated 24 designs by Latin hypercube sampling over seven inputs. The fiber-resolved model used 137 chains of two-node truss elements tied to the solid elements in the wall, with the solid stiffness reduced by the stiffness assigned to the trusses.

The light-responsive tip was a 400\,mm tube of inner radius 20\,mm and outer radius 40\,mm, with a Young's modulus of 9\,MPa and a Poisson's ratio of 0.45. The base was clamped and geometric nonlinearity was included. Illumination was modeled as axial contraction with a nominal amplitude of $1.0\%$ to $5.4\%$, scaled by side and axial position over eight bands. Relative fiber travel was measured from the difference in arc-length changes between channels 30\,mm to either side of the neutral axis. The whole-stem model joined the light-responsive tip to an 800\,mm graded base, with a rigid cap anchoring the sliding fibers. A 100\,N transverse load was applied at the junction of the tip and base. For the cantilever comparisons, a 100\,N transverse tip load was applied to the graded tube and to a uniform tube of equal area-mean modulus. Supplementary Methods, section~3, gives the loading directions, mesh limits, and displacement and stiffness measurements.

\section*{Code and data availability}

Code and data are available through the GitHub repository at \url{https://github.com/lamm-mit/FormalGen}.

\section*{Funding}
This work was funded by DARPA (COSMIC) and the MIT Generative AI Consortium (MGAIC).
Part of this work was supported by the U.S. Department of Energy, Office of Science, Office of Advanced Scientific
Computing Research and Office of Basic Energy Sciences, Scientific Discovery through Advanced Computing (SciDAC)
program under the FORUM-AI project.  

\section*{Author contributions}

L.M. and M.J.B. jointly conceived the study.
L.M. developed the framework, implemented the software, and performed the formalization and computational analyses.
Both authors interpreted the results.
L.M. prepared the figures and drafted the manuscript.
M.J.B. supervised the research and revised the manuscript.

\section*{Competing interests}

The authors declare no competing financial interests or personal relationships that could have influenced this work.

\bibliographystyle{naturemag}
\bibliography{export}

\includepdf[pages=-]{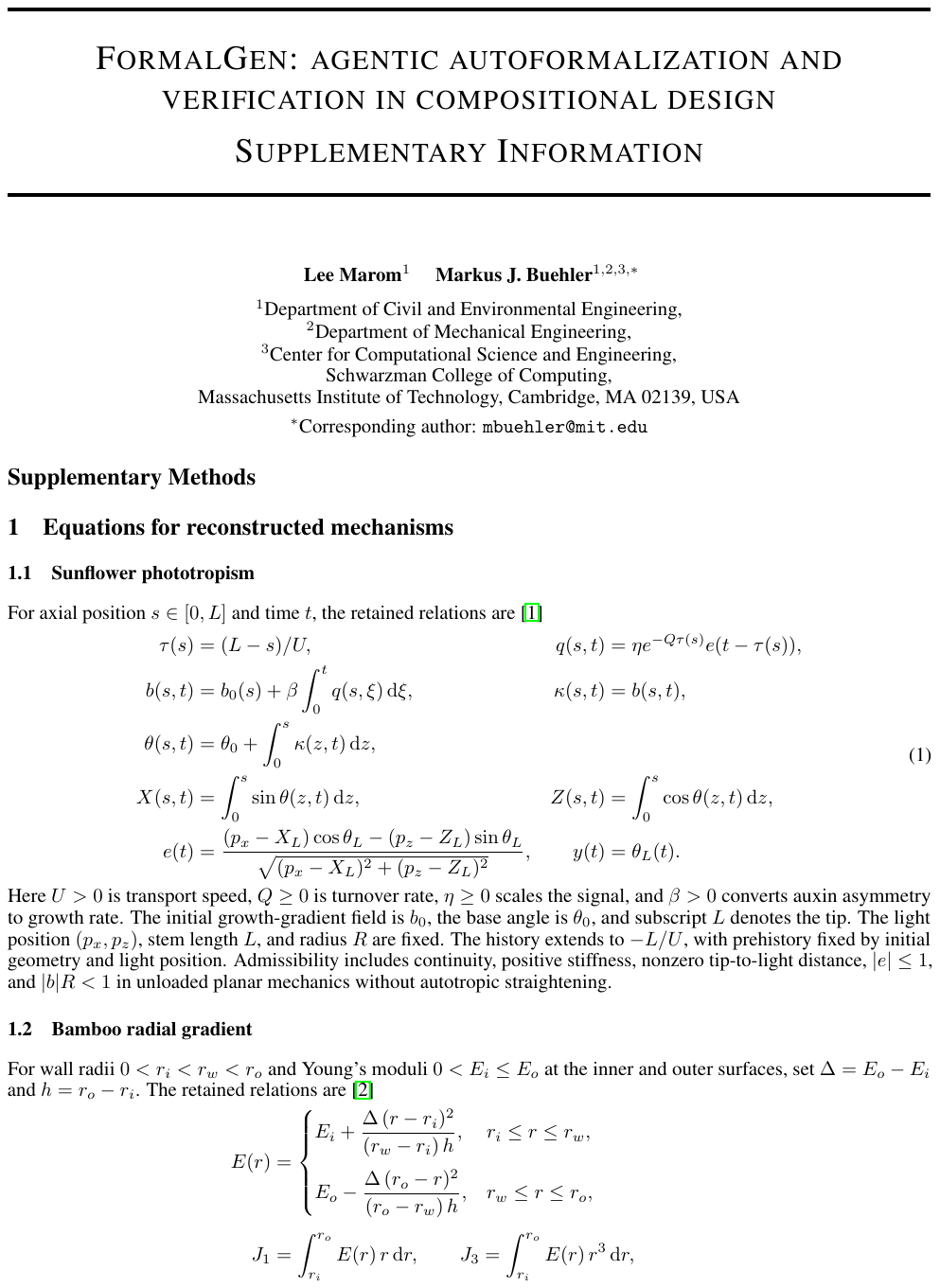}

\end{document}